\documentclass[runningheads]{llncs}
\usepackage{graphicx}

\usepackage{tikz}
\usepackage{comment}
\usepackage{amsmath,amssymb} 
\usepackage{color}

\usepackage[accsupp]{axessibility}  

\usepackage{epsfig}
\usepackage{booktabs}
\usepackage{multirow}
\usepackage{array}
\usepackage{tabularx}
\newcolumntype{P}[1]{>{\centering\arraybackslash}p{#1}}
\usepackage[pagebackref,breaklinks,colorlinks]{hyperref}
\usepackage[capitalize]{cleveref}

\crefname{section}{Sec.}{Secs.}
\Crefname{section}{Section}{Sections}
\Crefname{table}{Table}{Tables}
\crefname{table}{Tab.}{Tabs.}

\usepackage{float}
\usepackage{tikz}
\usepackage{comment}
\usepackage[pagebackref,breaklinks,colorlinks]{hyperref}
\usepackage{orcidlink}

\begin{document}
\pagestyle{headings}
\mainmatter
\def\ECCVSubNumber{1737}  



\title{A Reliable Online Method for Joint Estimation of Focal Length and Camera Rotation
}

\titlerunning{Reliable Online Estimation of Focal Length and Camera Rotation}
%
\author{Yiming Qian\inst{1}\orcidlink{0000-0002-1795-2038}  \and
	James H. Elder\inst{2}}
\authorrunning{Qian Y.\& Elder J.}
%
\institute{A*star, Institute of High Performance Computing, Singapore \email{qian\_yiming@ihpc.a-star.edu.sg} \and
	York University, Centre for Vision Research, Canada\\
	\email{jelder@yorku.ca}}

\maketitle

\begin{abstract}
Linear perspective cues deriving from regularities of the built environment can be used to recalibrate both intrinsic and extrinsic camera parameters online,
but these estimates can be unreliable due to irregularities in the scene, uncertainties in line segment estimation and background clutter.
Here we address this challenge through four initiatives.  First, we  use the PanoContext panoramic image dataset~\cite{zhang2014panocontext} to curate a novel and realistic dataset of planar projections over a broad range of scenes, focal lengths and camera poses.  
Second, we use this novel dataset and the YorkUrbanDB~\cite{denis2008efficient}  to systematically evaluate the linear perspective deviation measures frequently found in the literature and show that  the choice of deviation measure and likelihood model has a huge impact on reliability.  Third, we use these findings to create a novel system for online camera calibration we call $f\mathbf R$, and show that it outperforms the prior state of the art, substantially reducing error in estimated camera rotation and focal length.    Our fourth contribution is a novel and efficient approach to estimating uncertainty that can dramatically improve online reliability for performance-critical applications by strategically selecting which frames to use for recalibration.  
\end{abstract}

\section{Introduction}
Online camera calibration is an essential task for applications such as traffic analytics,  mobile robotics, architectural metrology and sports videography.  While intrinsic parameters can be estimated in the lab, these parameters drift due to mechanical fluctuations.  Online estimation of extrinsic camera parameters is crucial for translating observations made in the image to quantitative inferences about the 3D scene.  If a fixed camera is employed, extrinsics can potentially be estimated manually at deployment, but again there will be drift due to mechanical and thermal variations, vibrations and wind, for outdoor applications.  With longer viewing distances, rotational drift can lead to major errors that may be devastating for performance-critical tasks, such as judging the 3D distance between a pedestrian and a car.  For PTZ cameras and mobile applications, camera rotation varies over time.  PTZ encoder readings are not always available and become less reliable over time~\cite{wu2013keeping}, and for mobile applications IMU data are subject to drift.  For all of these reasons,  reliable visual methods for online geometric recalibration of camera intrinsics and extrinsics are important.  

\begin{figure*}[htpb!] 
\centering
{\includegraphics[width=0.75\columnwidth]{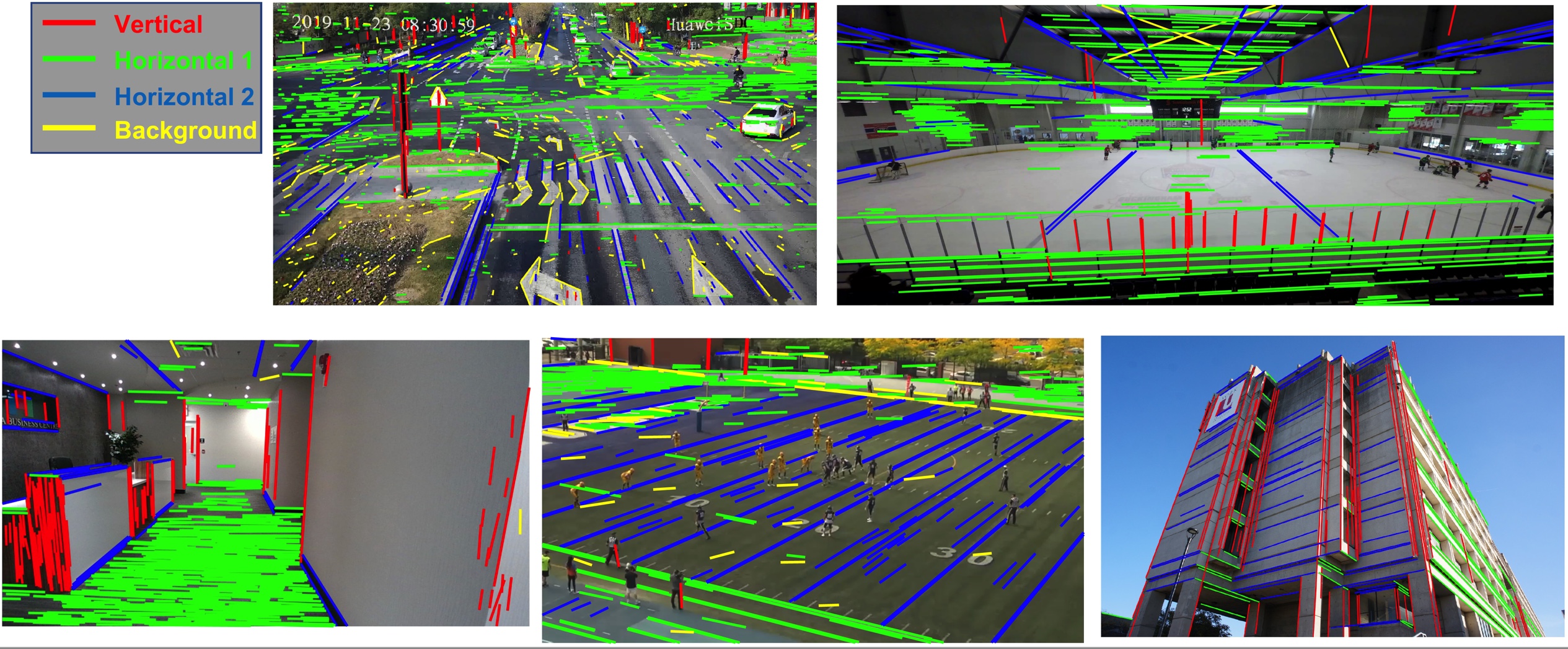}} 
\caption{ Example application domains of online camera calibration.  We show here labelling of line segments according to Manhattan direction, as determined by our novel {\em f}{\bf R} system that simultaneously estimates focal length and camera orientation.}
\label{fig:examples}
\end{figure*}

Coughlan and Yuille~\cite{coughlan1999manhattan} introduced an approach to online estimation of 3D camera rotation based on a ``Manhattan World" (3-point perspective) assumption, i.e., that a substantial portion of the linear structure in the image projects from three mutually orthogonal directions:  one vertical and two horizontal.  This assumption can apply quite generally in the built environment and is relevant for  diverse application domains including traffic analytics, mobile robotics, architecture and sports videography (Fig. \ref{fig:examples}).

Subsequent work, reviewed below, has built on this idea to improve accuracy and incorporate simultaneous estimation of focal length.  However, we have found that these methods vary a lot in accuracy depending upon the scene and whether focal length is known or estimated.  This motivates our attempt to  better understand how these  methods generalize to diverse scenes and knowledge of intrinsics, and to develop methods to estimate uncertainty of individual estimates.  


We address these limitations through four contributions: {\bf 1)} From the PanoContext dataset~\cite{zhang2014panocontext} we curate a novel and realistic PanoContext-{\em f}{\bf R} dataset of planar projections over a broad range of scenes, focal lengths and camera poses that can be used to evaluate systems for simultaneous estimation of focal length and camera rotation (tilt/roll).  
{\bf 2)} We use this novel dataset and the YorkUrbanDB~\cite{denis2008efficient}  to  evaluate the linear perspective deviation measures found in the literature and  show that  a) the choice of deviation measure has a huge impact on reliability, and b)  it is critical to also employ the correct likelihood models. 
{\bf 3)} We use these findings to create a novel system for online camera calibration we call $f\mathbf R$, and show that it  outperforms the  state of the art.
{\bf 4)} We develop a novel approach to estimating uncertainty in parameter estimates.
This is important for two reasons.  First, knowledge of uncertainty can be factored into risk models employed in engineering applications, co-determining actions designed to mitigate this risk.  Second, especially for cases where parameters are expected to be slowly varying, systems can sparsely select frames that are more likely to generate trustable estimates of camera parameters.  The source code and PanoContext-{\em f}{\bf R} dataset are available on GitHub\footnote{https://github.com/ElderLab-York-University/OnlinefR}.

\section{Prior work}
\label{sec:prior}
\subsection{Image features and deviation measures}
To apply linear perspective cues to camera calibration, two key design choices are 1) what features of the image to use and 2)  how to measure the agreement of these features with hypothesized vanishing points.  
Coughlan and Yuille's original approach~\cite{coughlan1999manhattan} employed a likelihood measure on the angular deviation between the line connecting a pixel to the hypothesized vanishing point (which we will refer to in the following as the {\em vanishing line} and the ray passing through the pixel in the direction orthogonal to the local luminance gradient.

\begin{figure}[htpb!] 
\begin{tabular}{ccccc}
	\includegraphics[width=0.19\textwidth]{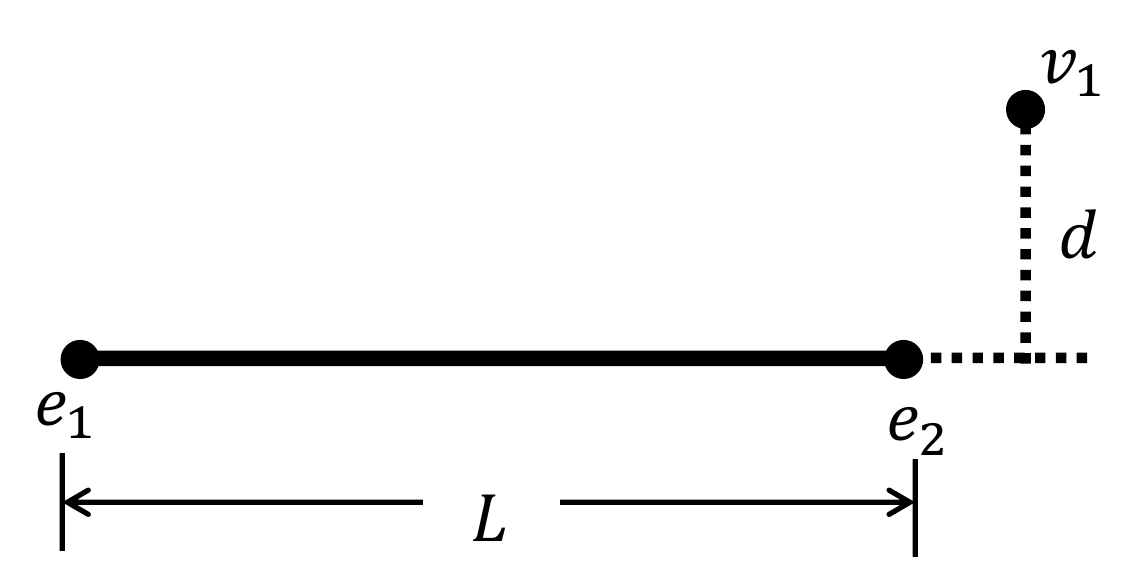}  &
	{\includegraphics[width=0.19\textwidth]{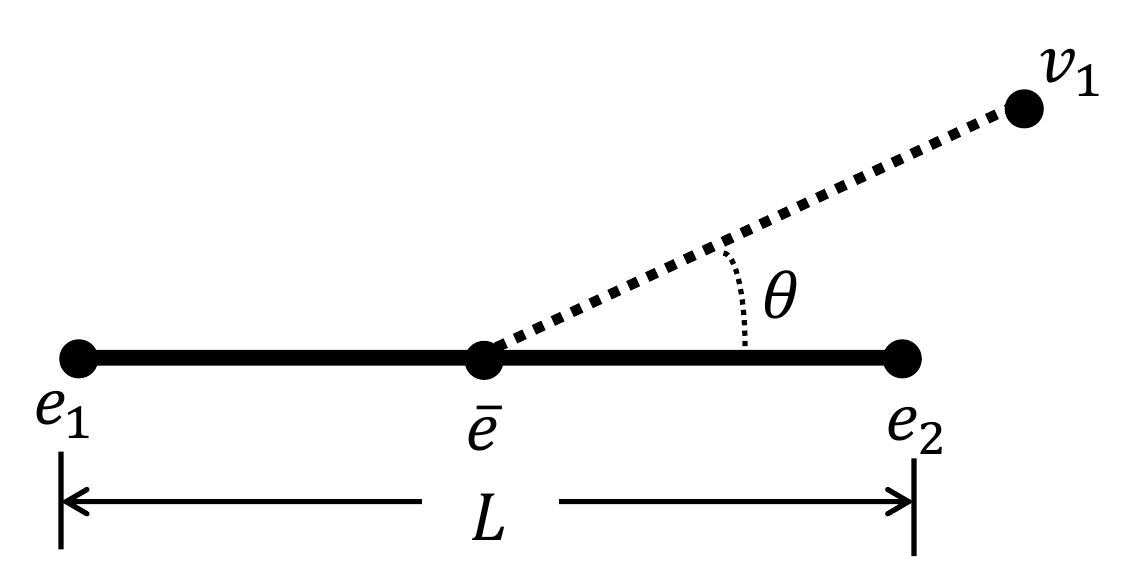}} &
	{\includegraphics[width=0.19\textwidth]{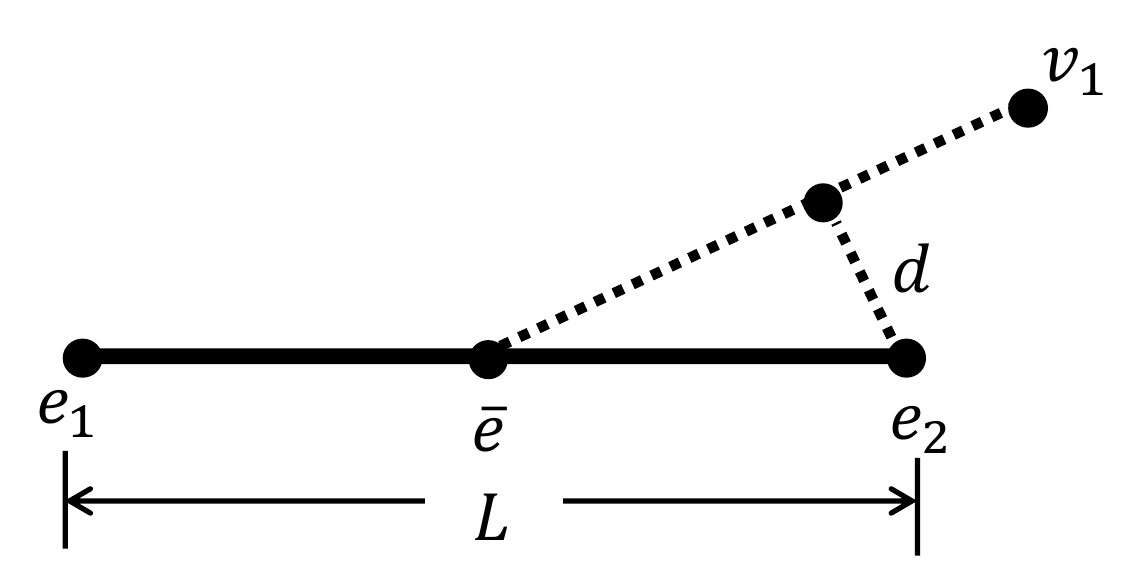}}  &
	{\includegraphics[width=0.19\textwidth]{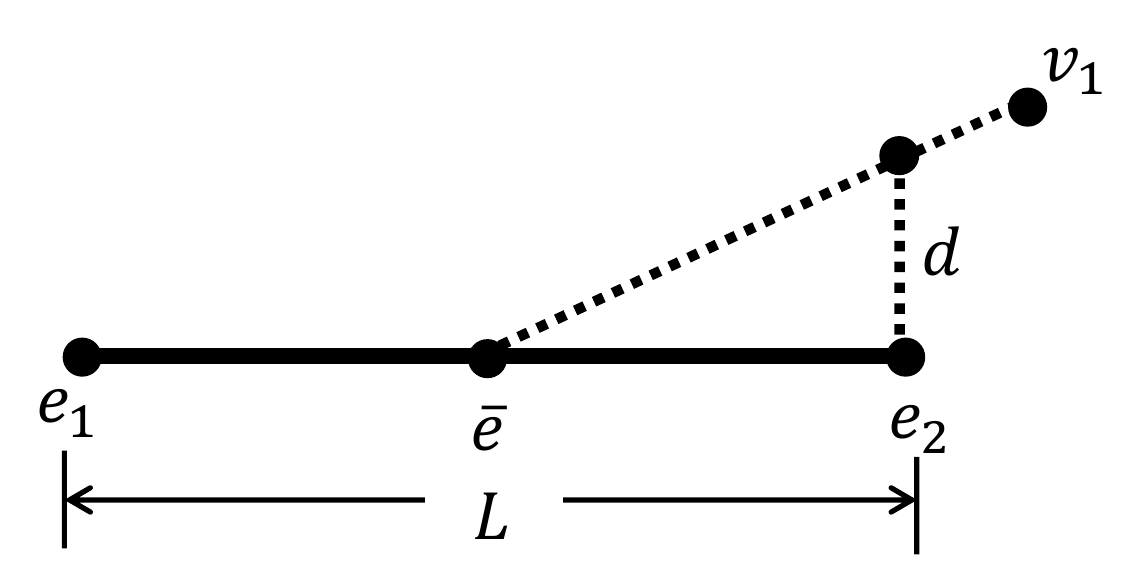}} &
	{\includegraphics[width=0.19\textwidth]{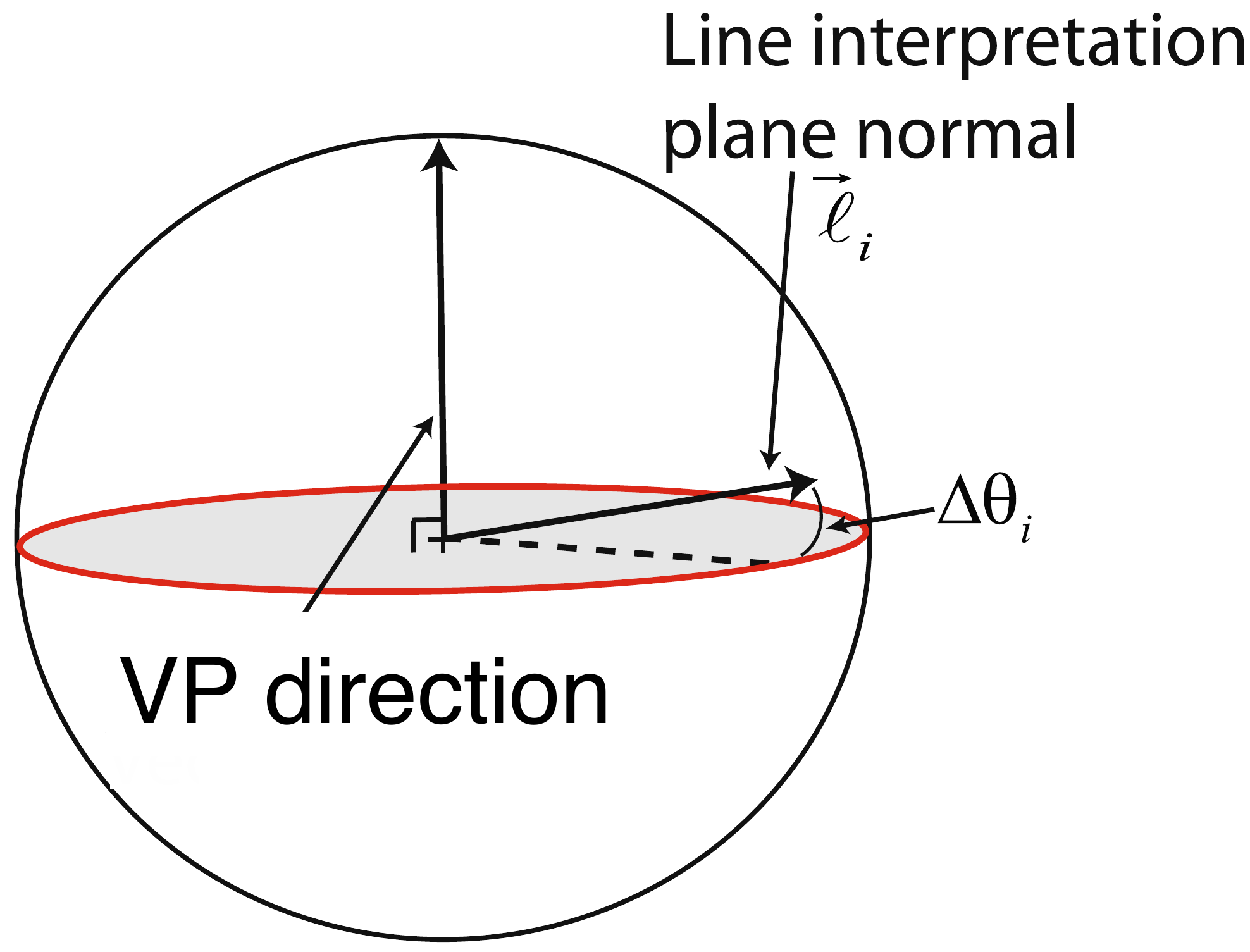}} \\
	(a) & (b) & (c) & (d) & (e)
\end{tabular}
\caption{ Five deviation measures evaluated (see also \cite{xu2013minimum}):  (a) distance from vanishing point to line passing through line segment. (b) angular deviation between line segment and the vanishing line through the line segment midpoint. (c) distance from line segment endpoint to vanishing line  (d)  distance from line segment endpoint to vanishing line, measured orthogonal to line segment.   (e) angle  between the interpretation plane normal and the plane orthogonal to the vanishing direction \cite{Tal:12}. }
\label{fig:method_distance}
\end{figure}

There have since been diverse attempts to improve on this approach, in large part by using different features and different measures of agreement.   
Deutscher et al.~\cite{deutscher2002automatic} and Ko{\v{s}}eck\`a and Zhang~ \cite{videocompass} first generalized this approach to allow simultaneous estimation of focal length.  While Deutscher et al. retained the gradient field representation of Coughlan and Yuille, in their Video Compass system, Ko{\v{s}}eck\`a and Zhang switched to using sparser line segments, and proposed a Gaussian likelihood model based upon the  projection distance from  the line passing through a line segment to its associated vanishing point (Fig. \ref{fig:method_distance} (a)).  

\setlength{\tabcolsep}{0.25em}

Rother~ \cite{rother2002new} used line segments as well, employing a heuristic version of the deviation measure similar to Coughlan and Yuille's (Fig. \ref{fig:method_distance} (b)).   Lee et al.~\cite{2009geometric} followed a similar approach to simultaneous estimation of focal length and camera rotation, and have open-sourced their code, allowing us to evaluate their system below.  Denis et al.~\cite{denis2008efficient} employed the same angular deviation measure for edges, but returned to Coughlan and Yuille's probabilistic framework, using a learned generalized exponential distribution to model these deviations.  

Tardif~\cite{tardif2009non} continued the practice of using line segments, but proposed  an alternative deviation measure based on the squared distance between segment endpoints and the vanishing line  (Fig. \ref{fig:method_distance} (c)).   Wildenauer~\cite{wildenauer2012robust} later adopted  the same measure but, similar to Denis et al.~\cite{denis2008efficient}, returned to Coughlan and Yuille's probabilistic framework, using a Cauchy distribution to model the signed distance of segment endpoints to the vanishing line.  

These three deviation measures were reviewed by Xu et al.~\cite{xu2013minimum}, who used intuitive arguments to suggest that each of these measures is flawed.  Based upon their analysis, they proposed a novel deviation measure that also uses the distance of the segment endpoints to the vanishing line, but measures this distance orthogonal to the {\em segment}, instead of orthogonal to the vanishing line (Fig. \ref{fig:method_distance} (d)).   Their full probabilistic model also favours vanishing points that lie far from the segment midpoint, along the segment direction.

Hough maps can be used to accurately detect lines in an image.  Motivated by the early work  of Collins \& Weiss~\cite{collins1990vanishing}, Tal \& Elder~\cite{Tal:12} proposed a probabilistic Hough method to estimate lines and then estimated  camera rotation based on a probabilistic model of deviation in the Gauss sphere, specifically, the angular deviation between the interpretation plane normal and the plane orthogonal to the vanishing direction (Fig. \ref{fig:method_distance}(e)).  

\subsection{Benchmarks}
The most prevalent dataset used for Manhattan scene analysis is the YorkUrbanDB dataset~\cite{denis2008efficient}, which does sample a broad range of scenes but is limited in that focal length is fixed and roll and tilt vary over a relatively modest range. Also, the 101 images in the dataset are too few to train a deep network.  More recently there have been efforts~\cite{hold2018perceptual,lee2020neural}  to create datasets large enough to train deep networks by curating random planar projections from existing panoramic image datasets such as SUN360~\cite{xiao2012recognizing} and Google StreetView.  This makes it easy to generate enough images to train a deep network, and also ensures that we have exact ground truth for camera focal length, tilt and roll angle.  Unfortunately, there is no ground truth for camera pan, and so networks trained on these datasets are unable to recover the full camera rotation ${\bf R}$.


\subsection{State-of-the-Art Systems}
In this paper we are focused on the problem of online estimation of focal length $f$ and camera rotation ${\bf R}$.   Unfortunately, many of the systems reviewed above assumed known focal length, and/or did not open-source their code to allow comparison.  
The exception, as mentioned, is the line segment method of Lee et al.~\cite{2009geometric} (Fig. \ref{fig:method_distance} (b)) which we compare against below.  

More recently, Simon et al.~\cite{simon2016simple} use the angular deviation measure in Fig. \ref{fig:method_distance}(b) but introduce a preprocessing step to estimate the horizon as a guide to estimating Manhattan vanishing points.  They have open-sourced their code.

Hold-Geoffroy et al.~\cite{hold2018perceptual} kicked off the use of deep learning for online camera calibration, using the Sun360 panoramic image dataset~\cite{xiao2012recognizing} to generate planar projections based on random samplings of focal length, horizon, roll and camera aspect ratio.  They used this dataset to train, validate and test a network based on
a denseNet  backbone and three separate  heads to estimate the horizon height and angle and the vertical FOV, from which camera focal length, roll and tilt can be derived.  While the Hold-Geoffroy et al.~\cite{hold2018perceptual} dataset has not been released and the method has not been open-sourced, they do provide a web interface for testing their method on user-provided images, allowing us to evaluate and compare their approach on public datasets.

Lee et al.~\cite{lee2020neural} followed a similar approach to generating large datasets (see above) to train a {\em neural geometric parser}.  Their system takes  line segments detected by LSD~\cite{von2008lsd} as input, first estimating the zenith vanishing point, selecting segments that are close to vertical, generating vanishing point hypotheses from pairs of these segments, and then feeding the segments and the hypotheses into a deep network to score the hypotheses.  High-scoring zenith candidates are then used to generate camera tilt/roll and focal length hypotheses that are combined with the image and Manhattan line maps as input to  a second network to generate final camera tilt/roll and focal length estimates.  In a more recent paper, Lee et al.~\cite{lee2021ctrl} have introduced an updated system called CTRL-C that continues to employ LSD line segments but shifts to a transformer architecture for estimating the parameters.  While the neural geometric parser has not been open-sourced, CTRL-C has, and we compare with their approach below.

There are other recent deep learning approaches to estimation of  vanishing points and/or camera rotation~\cite{xian2019uprightnet,li2019quasi,li2021learning}, but these do not estimate focal length.   

\section{$f\mathbf R$ Method}
To develop our $f\mathbf R$ method for online camera calibration, 
we will assume a camera-centred world frame aligned with the Manhattan structure of the scene and employ a standard projection model 
$
\tilde{\mathbf x} = \mathbf {K R}\bar{\mathbf X}
$
where $\tilde{\mathbf x}$ is an image point in homogenous coordinates, ${\mathbf K}$ is the camera's intrinsic  matrix, ${\mathbf R}$ is the camera rotation matrix and $\bar{\mathbf X}$ is a 3D world point in augmented coordinates.  
Unless otherwise noted, we will assume that any nonlinear distortions in the camera have been calibrated out in the lab, and that after laboratory calibration the camera has a central principal point, square pixels and zero skew, leaving a single intrinsic unknown, the focal length $f$.  While these assumptions will not be met exactly by real cameras in the field, they are the standard assumptions employed by all of the methods reviewed above and the methods we compare against below, and in our experience are reasonable approximations for  most higher-quality modern cameras.  

The goal of $f\mathbf R$ then is to estimate the unknown focal length $f$, together with the 3D rotation ${\bf R}$ of the camera relative to the Manhattan frame of the scene.  Again, we note that this goes beyond the capacity of recent deep learning approaches, which are only able to estimate two of the three rotational degrees of freedom (roll and tilt).   Standard formulae can be used to convert between camera Euler angles (roll, tilt, pan) and  the camera rotation matrix $\mathbf R$~\cite{gaugeras1993three}.   

We collect these unknowns into the parameter set $\Psi = \{f,{\bf R}\}$.  Note that $\Psi$ completely determines the locations of the three Manhattan vanishing points:  The $i$th vanishing point is given by $\tilde{\mathbf x}_i = \mathbf {KR}_i$, where ${\mathbf R}_i$ is the $i$th column of ${\mathbf R}$.  

In order to better understand the relative advantages of the various deviation measures proposed in the literature, and the role of probabilistic modeling, we will adopt the original mixture model framework first proposed by Coughlan \& Yuille~\cite{coughlan1999manhattan} for luminance gradients and later extended to line segments~ \cite{videocompass}, edges~\cite{denis2008efficient} and  lines~\cite{Tal:12}.  We will focus on the use of line segments as features, as there are numerous high-quality open-source line segment detectors now available~\cite{von2008lsd,almazan2017mcmlsd,lin2020deep} and line segments have been shown to be effective features for Manhattan scene analysis.  We will use the MCMLSD line segment detector~\cite{almazan2017mcmlsd} for most of the experiments below, but will also evaluate the advantages and disadvantages of alternative detectors.

\subsection{Probabilistic model}
$f\mathbf R$ will assume that each line segment $\vec{l}_i$ in the image is generated by a model $m_i\in M$ corresponding to one of four possible processes:  the three Manhattan families of parallel lines or a background process assumed to generate a uniform distribution of lines.  Given camera parameters $\Psi$, the probability of observing the segment $\vec{l}_i$ is then given by:
\begin{equation}
p(\vec{l}_i|\Psi )=\sum_{m_i\in M} p(\vec{l}_i|\Psi,m_i) p(m_i)
\label{eqn:mixture}
\end{equation}
where $p(m_i)$ is the prior probability that a segment is generated by process $m_i$ and $p(\vec{l}_i|\Psi,m_i)$ is the likelihood of an observed line generated by $m_i$ under camera parameters $\Psi$.

Since, as noted above, the parameters $\Psi$ completely determine the vanishing points, the likelihoods $p(\vec{l}_i|\Psi,m_i)$ for Manhattan processes $m_i$ can be parameterized by one of the non-negative measures of deviation between the line $\vec{l}_i$ and the corresponding Manhattan vanishing point (deviation measures {\bf a-e} reviewed above and summarized in Fig. \ref{fig:method_distance}).  We will generally use the YorkUrbanDB training partition to fit exponential models $p(x)=\frac{1}{\lambda}\exp(-x/\lambda)$ for these deviations, and assume that the background process is uniform on this measure - see supplemental material for the fits.  The one exception is for deviation measure d, where we employ the central Gaussian model proposed  by Xu et al.~\cite{xu2013minimum}, which also favours vanishing points that lie far from the segment midpoint, along the segment direction.  Here  the dispersion parameter $\sigma$ does not have a clear interpretation in terms of the YorkUrbanDB so we use the value of $\sigma=1$ pixel recommended by Xu et al.  Table  \ref{tab:parametersEXP} summarizes these parameters.

\begin{table}[htpb!] 
\caption{Dispersion parameters for likelihood models.}
\vspace{-1em}
\centering
\begin{tabular}{|c|c|c|r|r|}
	\hline
	\textbf{Deviation measure} & \textbf{Model} & \textbf{Param}       & \textbf{Horiz} & \textbf{Vert} \\ \hline
	a             & Exponential & $\lambda$ & 94.46 pix                     & 17.26 pix                  \\ \hline
	b             & Exponential & $\lambda$ & 1.46 deg                      & 0.57 deg                    \\ \hline
	c             & Exponential & $\lambda$ & 0.39 pix                    & 0.53 pix                 \\ \hline
	d             & Gaussian    & $\sigma$  & 1.00 pix                            & 1.00 pix                          \\ \hline
	e             & Exponential & $\lambda$ & 0.80 deg                       & 0.57 deg                   \\ \hline
\end{tabular}
\label{tab:parametersEXP}
\end{table}

The priors $p(m_i)$ are also estimated from the YorkUrbanDB training set: 45\% of lines are expected to arise from the vertical process, 26\% from each of the horizontal processes and 3\% from the background process.


Assuming conditional independence over $n$ observed line segments $\vec{l}_i$ and a flat prior over the camera parameters $\Psi$, the optimal solution should be found by maximizing the sum of log likelihoods, however we find empirically that we obtain slightly better results if we weight by line segment length $\left |\vec{l}_i \right |$, solving for 
\begin{equation}
\Psi^*=\arg\max_\Psi \sum_{i=1,\ldots,n} \left |\vec{l}_i \right |\log p\left(\vec{l}_i|\Psi \right)
\label{eqn:psistar}
\end{equation}
We conjecture that the small improvement achieved by weighting by line length may derive from a statistical dependence between line length and the likelihood model:  Longer lines may be more accurate. 

\subsection{Parameter search}
\label{sec:search}
The $f\mathbf R$ objective function (Eqn. \ref{eqn:psistar}) is generally non-convex.  We opt for a simple search method that can be used to compare the deviation measures {\bf a-e}  summarized in Fig. \ref{fig:method_distance} and to assess the role of probabilistic modeling.  The method proceeds in two stages.  In Stage 1, we do a coarse $k\times k\times k\times k$ grid search over the four-dimensional parameter space of $\Psi$, evaluating  Eqn. \ref{eqn:psistar} at each of the $k^4$ parameter proposals.  Specifically, we sample uniformly over the range $[-45,+45]$ deg for pan, $[-15,+15]$ deg for roll, $[-35,+35]$ deg for tilt and $[50,130]$ deg for horizontal FOV.  (We also evaluated a  RANSAC approach ~\cite{wildenauer2012robust} for Stage 1 but found it be less accurate - see supplementary material for results.)  We then deploy a nonlinear iterative search (MATLAB fmincon) initialized at each of the the top $l$ of these $k^4$ proposals and constrained to solutions within the ranges above.  In the following we will use $k=8, l=4$.

Note that focal length $f$ and FOV are monotonically related through ${\rm FOV} = 2\arctan\left(\frac{w}{2f}\right)$, where $w$ is the image width.  While we search over FOV we will generally convert to focal length when reporting results.

\subsection{Error prediction}
\label{sec:esterr}
One advantage of a probabilistic approach to online camera calibration is the opportunity to predict when estimates can be trusted.  Since we explicitly model the distribution of deviations between line segments and hypothesized vanishing points, we might hope to estimate uncertainty in camera parameters using standard error propagation methods.  
Unfortunately, we find empirically that this local method does not lead to very accurate estimates of uncertainty, perhaps because the derivative of the camera parameters with respect to the segment orientations at the estimated parameters $\hat\psi$ is not very predictive of the derivative at the true camera parameters $\psi^*$ (see supplementary material for details).  We therefore propose instead three easily computable global predictors to inform the level of trust that should be invested in estimated camera parameters:

1) {\bf Number of line segments.}  We assign each detected segment$\vec{l}_i$ to the most  likely generating process $m_i$ under  our mixture model (Eqn. \ref{eqn:mixture}), count those assigned to each Manhattan direction, and take the minimum, to reflect the intuition that this will be the weakest link in the global parameter estimation process.  
2) {\bf Entropy.}  This cue takes advantage of the first stage of our parameter search.  We conjecture that more reliable parameter estimation will be reflected in a more peaked distribution of likelihoods over the $k^4$ parameter proposals evaluated.  To capture this intuition, we normalize the likelihoods into a discrete probability distribution and compute its entropy.  Low entropy distributions are expected to be more reliable.
3) {\bf Likelihood.}  We compute the  mean  log likelihood of the final estimate output from the second nonlinear iterative stage of our search, normalized by the number of line segments.

To predict camera parameter MAE from these  cues we fit a KNN regression model on the PanoContext-$fR$ training partition - see Section \ref{sec:esterrresults} for details.

\section{Datasets}
We evaluate $f\mathbf R$ and competing methods on two datasets:  the YorkUrbanDB~\cite{denis2008efficient} test partition, and our novel PanoContext-{\em f}{\bf R} dataset, curated from \cite{zhang2014panocontext}.

The YorkUrbanDB test partition consists of 51 indoor and outdoor images.  Camera focal length $f$ was fixed and estimated in the lab.  Manhattan line segments were hand-labelled to allow estimation of the camera rotation $\bf R$~\cite{denis2008efficient}.  While the camera was handheld, the variation in camera roll and tilt is fairly modest:   standard deviation of 0.83 deg for roll and 4.96 deg for tilt.  

To allow evaluation over a wider range of camera parameters and  to assess how well recent deep networks generalize compared to geometry-driven approaches, we introduce  PanoContext-{\em f}{\bf R}, a curation of planar projections from the PanoContext dataset~\cite{zhang2014panocontext}, which contains 706 indoor panoramic scenes.  We randomly divide this dataset into two equal training and test partitions of 353 scenes each.  We will use the test partition to evaluate and compare deviation measures, probabilistic models and state-of-the-art systems for camera parameter estimation.  While our {\em f}{\bf R} system is not a machine learning method and does not require the training partition, we will explore methods for learning to predict estimation error in Section \ref{sec:esterr} and will use the training dataset there. 

\begin{figure}[htpb!] 
\centering
{\includegraphics[width=0.4\columnwidth]{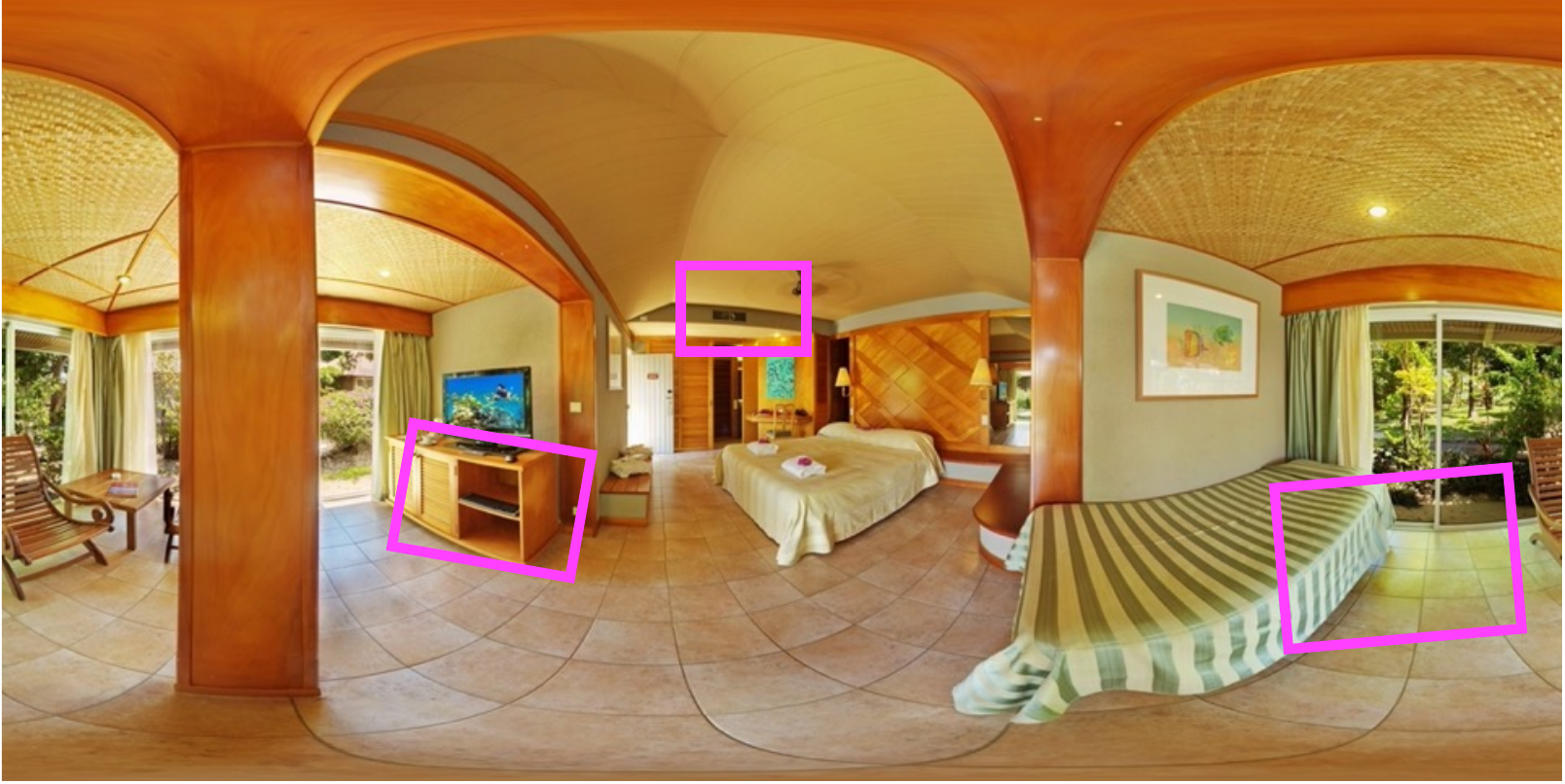}} 
\caption{Three example planar projections from our new PanoContext-{\em f}{\bf R} dataset. }
\label{fig:samples}
\end{figure}

For each of these scenes we generated 15 planar projections, using a standard  $640\times480$ pixel resolution.  
Three images were sampled for each of 5 fixed horizontal  FOVs from 60 to 120 deg in steps of 15 deg.  Pan, roll and tilt were randomly and uniformly sampled over [-180,+180], [-10,+10] and [-30,+30] ranges respectively (Fig. \ref{fig:samples}).  This generated 5,295 images for each of the training and test partitions.
This dataset is not intended for network training, but rather for  evaluating generalization of pre-trained and geometry-driven models.

We sampled FOVs discretely for  PanoContext-{\em f}{\bf R} so we could clearly see whether algorithms are able to estimate focal length (see below).
However, we have also created and will make publicly available an alternative PanoContext-u{\em f}{\bf R} that samples horizontal FOV randomly and uniformly over [60,120] deg. (See supplementary material.)
Since the the YorkUrbanDB provides the ground truth rotation matrix, we can evaluate the frame angle error, i.e., the magnitude of the rotation required to align the 
estimated camera frame with the ground truth frame.  Since for PanoContext-{\em f}{\bf R} we do not have ground truth pan, we report mean absolute roll and tilt errors.

\section{Experiments}
\subsection{Evaluating deviation measures}
We begin by applying the $f\mathbf R$ search method (Section \ref{sec:search}) to identify the parameters $\Psi$ maximizing agreement between line segments and vanishing points, based on the five deviation measures {\bf a-e}  summarized in Fig. \ref{fig:method_distance}.  Without probabilistic modeling, we employed the objective function  $\sum_i\log\delta_i$, where $\delta_i=d_i,\theta_i,\Delta\theta_i$ depending on which of the five deviation measures is employed.   (We found empirically that logging the deviations before summing improved results.).  We found that none of the deviation measures yielded good results:  Mean focal length error remained above 11\% on the York Urban DB and above 34\% on the PanoContext-{\em f}{\bf R} dataset (See Supplementary Material for detailed results.)

Probabilistic modeling of the deviation measure greatly improves results (see Table \ref{tab:panoramicMethods} for results on the PanoContext-{\em f}{\bf R} dataset and supplementary material for results on the YorkUrbanDB).  We note that this represents the learning of only 4 parameters (horizontal and vertical dispersions and priors).  Moreover this learning seems to generalize well, as the parameters used to evaluate on the PanoContext-{\em f}{\bf R} dataset were learned on the YorkUrbanDB training set. 


\begin{table}[htpb]
\caption{Evaluation of deviation measures on the PanoContext-{\em f}{\bf R} training set.  Numbers are mean$\pm$standard error. }
\centering

\begin{tabular}{|l|c|c|c|r|}
	\hline
	\textbf{Dev. measure} &\textbf{Roll MAE (deg)}&\textbf{Tilt MAE (deg)}& \textbf{Focal length MAE (\%)}\\ \hline 
	a   &4.53$\pm$0.045 &14.2$\pm$0.14&     35.6$\pm$0.35            \\\hline    
	{\bf b ($f\mathbf R$)}   &\textbf{0.78$\pm$0.02} &\textbf{1.59$\pm$0.06}&  \textbf{8.4$\pm$0.26}           \\\hline    
	c    &0.90$\pm$0.009 &2.26$\pm$0.022&   14.1$\pm$0.14             \\\hline    
	d   & 0.81$\pm$0.008&1.67$\pm$0.015&   10.1$\pm$0.08             \\\hline    
	e    & 0.89$\pm$0.009&2.13$\pm$0.021&   19.5$\pm$0.19              \\\hline     
\end{tabular}
\label{tab:panoramicMethods}
\end{table}

Our second observation is that performance also depends strongly on the deviation measure, even when carefully modeling the likelihood.  The large errors produced by measure {\bf a} likely reflect the fact that vanishing points are often far from the principal point, resulting generally in very unpredictable deviations from the line passing through an associated line segment.
The other measures  yield better performance, but we note that deviation measure {\bf e} tends to have higher errors for focal length.  We believe this is because the Gauss Sphere measure of deviation suffers from a degeneracy:  As focal length tends to infinity, the interpretation plane normals collapse to a great circle parallel to the image plane.  Thus a hypothesized vanishing point near the principal point  will always generate small deviations, leading to a large likelihood.    

This leaves deviation measures {\bf b-d}.  Method {\bf d} performs less well on the YorkUrbanDB dataset and method {\bf c} has higher errors on the PanoContext-{\em f}{\bf R} dataset.  Overall, method {\bf b},  the method originally used by Coughlan \& Yuille for isophotes~\cite{coughlan1999manhattan}, Rother for line segments~ \cite{rother2002new}  and Denis et al. for edges~\cite{denis2008efficient}, appears to generate the most consistent performance, and so we adopt this method as our standard $f\mathbf R$ deviation measure for the remainder of the paper.

\subsection{Evaluating line segment detectors}
\label{sec:LSD}
All of the experiments above employed the MCMLSD line segment detector~\cite{almazan2017mcmlsd}.  Table \ref{tab:LSDPano} assesses the sensitivity of our system to this choice by substituting two alternative detectors:  The well-known LSD detector~\cite{von2008lsd} and a more recent deep learning detector called HT-LCNN~\cite{lin2020deep}.  Performance on the PanoContext-{\em f}{\bf R} dataset is significantly better using MCMLSD than the more recent HT-LCNN, despite the fact that HT-LCNN is reported~\cite{lin2020deep} to have much better precision-recall performance on the YorkUrban DB and Wireframe~\cite{xu2015statistical} datasets.  

We believe this discrepancy stems largely from the incompleteness of these datasets, which do not claim to label {\em all} Manhattan line segments.  Methods like MCMLSD, which attempt to detect all line segments (Manhattan and non-Manhattan), thus tend to achieve lower precision scores on these datasets, while deep learning methods like HT-LCNN learn to detect only the labelled segments and thus achieve higher precision scores.  Our experiment reveals that this higher precision does not translate into better performance but rather {\em worse} performance, presumably because HT-LCNN fails to generate Manhattan segments that may be unlabelled but are nevertheless useful for estimating the camera parameters.  We proceed with the MCMLSD line segment detector but will return to this choice when considering run-time efficiency (Section \ref{sec:runtime}).


\begin{table}[htpb]
\caption{Evaluating the choice of line segment detector on the PanoContext-{\em f}{\bf R} training set.   Numbers are mean $\pm$ standard error.}
\centering
\begin{tabular}{|l|c|c|c|c|r|}
	\hline
	\textbf{Methods}       & {\bf Run time }              &\textbf{Roll MAE }&\textbf{Tilt MAE }& \textbf{Focal length MAE }\\   
	      &  {\bf (sec) }             &\textbf{(deg)}&\textbf{ (deg)}& \textbf{ (\%)}\\ \hline  
	MCMLSD~\cite{almazan2017mcmlsd}   & 4.23 &${\bf 0.78\pm0.02}$ &${\bf 1.59\pm0.06}$&  $\bf{8.4\pm0.26}$           \\\hline    
	LSD~\cite{von2008lsd}    & 0.40 &$1.23\pm0.012$ & $3.35\pm0.03$ &  $8.6\pm0.11$           \\\hline    
	HT-LCNN~\cite{lin2020deep} & 2.91 & $0.93\pm 0.009$&$1.91\pm 0.02$&    10.1$\pm$0.14         \\\hline    
\end{tabular}
\vspace{-1em}
\label{tab:LSDPano}
\end{table}

\subsection{Comparison with state of the art}
We compare our {\em f}{\bf R} system against the four state-of-the-art systems~\cite{2009geometric,simon2016simple,hold2018perceptual,lee2021ctrl} reviewed in Sec. \ref{sec:prior}.  Three of these~\cite{2009geometric,simon2016simple,lee2021ctrl}  are open-sourced and we downloaded the code.  The fourth (Hold-Geoffroy)~\cite{hold2018perceptual} provides a web interface.  We test two versions of the CTRL-C~\cite{lee2021ctrl} network:  CTRL-C S360 was trained on a dataset curated from the SUN360 dataset~\cite{xiao2012recognizing}.  CTRL-C GSV was trained on a dataset curated from the Google Street View dataset~\cite{lee2020neural}.

Table \ref{tab:SOTA})(top) shows results of this comparison on the York UrbanDB test set.  We report  roll and tilt error instead of  frame error, since the deep learning methods (Hold-Geoffroy~\cite{hold2018perceptual} and CTRL-C~\cite{lee2021ctrl}) do not estimate pan angle.   (See supplemental material for comparison of pan angle error with the methods of Lee et al.~\cite{2009geometric} and Simon et al.~\cite{simon2016simple}).  (The Simon et al. method returned a valid result for only 69\% of the York UrbanDB test images and 46\% of the PanoContext-$f \mathbf R$ test images;  we therefore report their average performance only for these.)

\begin{table}[htpb]
\caption{Performance comparison with SOA on the York UrbanDB (top) and PanoContext-{\em f}{\bf R} (bottom) test sets.  Numbers are mean $\pm$ standard error.}
\centering

\begin{tabular}{|l|c|c|c|c|c|}
	\hline
		\textbf{YorkUrbanDB}   & {\bf Run time}                  &\textbf{Roll }&\textbf{Tilt  }& \textbf{FOV  }&\textbf{Focal len.  }\\ 
		& {\bf (sec)}                  &\textbf{(MAE deg)}&\textbf{MAE (deg)}& \textbf{MAE (\%)}&\textbf{MAE (\%)}\\ \hline 
		Lee\cite{2009geometric}  & 1.13 & 0.80$\pm$0.2&1.33$\pm$0.2&     $9.2\pm1.15$      & $11.5\pm1.62$     \\\hline    
		Simon\cite{simon2016simple}  & 0.44 &2.63$\pm$0.6&11.0$\pm$1&   $11.4\pm2.56$    &  $15.1\pm2.08$    \\\hline    
		Hold-Geoffroy\cite{hold2018perceptual}  & n/a &1.47$\pm$0.2&7.57$\pm$0.6&   $48.7\pm2.11$ & $37.7\pm7.02$     \\\hline    
		CTRL-C\cite{lee2021ctrl} S360 & {\bf 0.32} &1.38$\pm$0.2&{\bf 1.09$\pm$0.1}& $29.0\pm2.50$  &$24.7\pm1.80$ \\\hline    
		CTRL-C \cite{lee2021ctrl} GSV & {\bf 0.32} &1.37$\pm$0.2&3.90$\pm$0.3& $69.4\pm1.66$   & $48.8\pm0.77$      \\\hline   
		{\em f}{\bf R}   & 6.40 &\textbf{0.50$\pm$0.1} &1.16$\pm$0.1&$\bf{3.8\pm0.61}$   & $\bf{4.6\pm0.78}$      \\\hline    
		Fast-{\em f}{\bf R}   & 0.89 & 1.13$\pm$0.4 &1.62$\pm$0.3&   $5.1\pm1.04$  &   $5.7\pm1.21$    \\\hline    
\end{tabular}
\vspace{1em}

	\begin{tabular}{|l|c|c|c|c|c|c|}
		\hline
		\textbf{PanoContext-{\em f}{\bf R}}                     &\textbf{Roll }&\textbf{Tilt }& \textbf{FOV }&\textbf{Focal len. }\\ 
		 &\textbf{MAE (deg)}&\textbf{MAE (deg)}& \textbf{MAE (\%)}& \textbf{(MAE \%)}\\ \hline 
		Lee\cite{2009geometric}  & 2.02$\pm$ 0.02&3.35$\pm$0.03&      $13.4\pm0.22$    &$22.3\pm0.22$      \\\hline    
		Simon\cite{simon2016simple}  &1.50$\pm$0.02&3.18$\pm$ 0.05&  $32.8\pm0.46$  &  $54.0\pm0.53$    \\\hline    
		Hold-Geoffroy\cite{hold2018perceptual}  & 1.58$\pm$0.02&3.69$\pm$0.04&  $16.1\pm0.12$  &  $20.5\pm0.20$   \\\hline    
		CTRL-C\cite{lee2021ctrl} S360 &1.03$\pm$0.01&2.19$\pm$0.02& $8.1\pm0.07$  & $13.1\pm0.13$         \\\hline    
		CTRL-C \cite{lee2021ctrl} GSV &2.24$\pm$0.02 &8.98$\pm$ 0.09& $17.3\pm0.11$ & $21.9\pm0.31$         \\\hline    
		{\em f}{\bf R }    &\textbf{0.78$\pm$0.02} &\textbf{1.59$\pm$0.06}&  \textbf{5.3$\pm$0.15} &   8.4$\pm$0.26    \\\hline    
		Fast-{\em f}{\bf R}    &0.89$\pm$0.02 &1.90$\pm$0.07&  $5.4\pm0.16$   &$\bf{7.6\pm0.23}$        \\\hline   
	\end{tabular}
\label{tab:SOTA}
\end{table}

All methods have fairly low error in roll, but remember that in the York UrbanDB, variation in camera roll is limited (standard deviation of 0.83 deg).  Our {\em f}{\bf R} system improves on the next best method (Lee et al.~\cite{2009geometric}) by 38\%.  
For tilt estimation, variability in performance across systems is more pronounced, but three systems (our {\em f}{\bf R} system, Lee et al.~\cite{2009geometric} and CTRL-C S360~\cite{lee2021ctrl}  all perform well.
Focal length estimation is the big differentiator.  Our {\em f}{\bf R} system performs much better than all of the other systems, beating the next best system (Lee et al.~\cite{2009geometric}) by 57\% (reducing MAE from 11.5\% to 4.9\%).
Notice that the deep learning methods (Hold-Geoffroy~\cite{hold2018perceptual} and CTRL-C~\cite{lee2021ctrl}) do not perform well.

Table \ref{tab:SOTA}(bottom) compares these systems on our novel PanoContext-{\em f}{\bf R} dataset.  (While FOV and focal length are monotonically related, we show MAE for both for the reader's convenience.)
We find that our {\em f}{\bf R} system performs significantly better that all of the other systems  on all three benchmarks (roll, tilt and focal length), improving on the next best system (CTRL-C S360~\cite{lee2021ctrl}) by 23\%, 25\% and 33\%, respectively.  (The relatively strong performance of the CTRL-C S360 system here may derive from the fact that the PanoContext dataset was original drawn from the SUN360 dataset on which CTRL-C S360 is trained.)

\begin{figure}[htpb!] 
	\centering
	\begin{tabular}[t]{cccc}
		{\includegraphics[width=0.125\columnwidth]{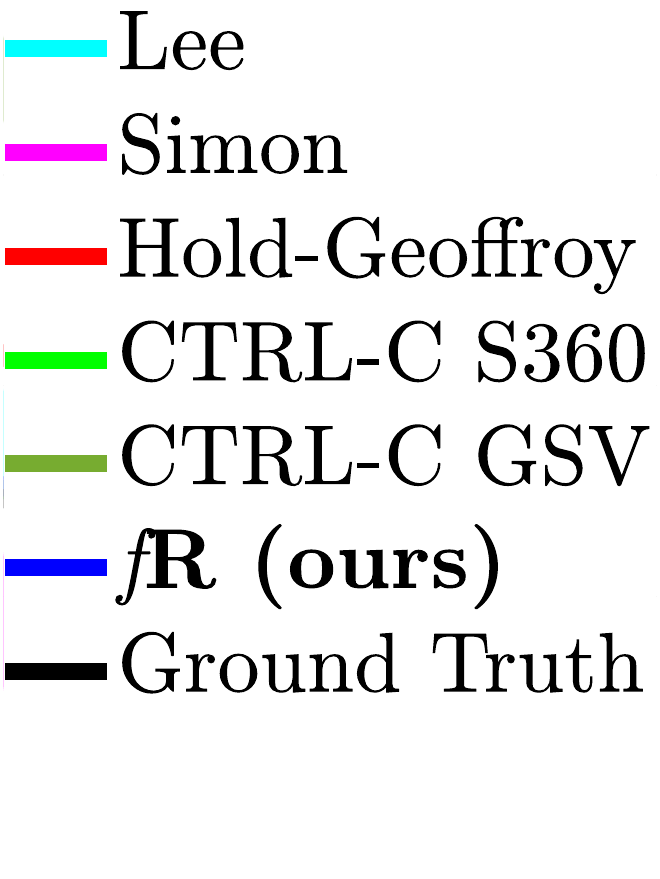}} &
		{\includegraphics[width=0.25\columnwidth]{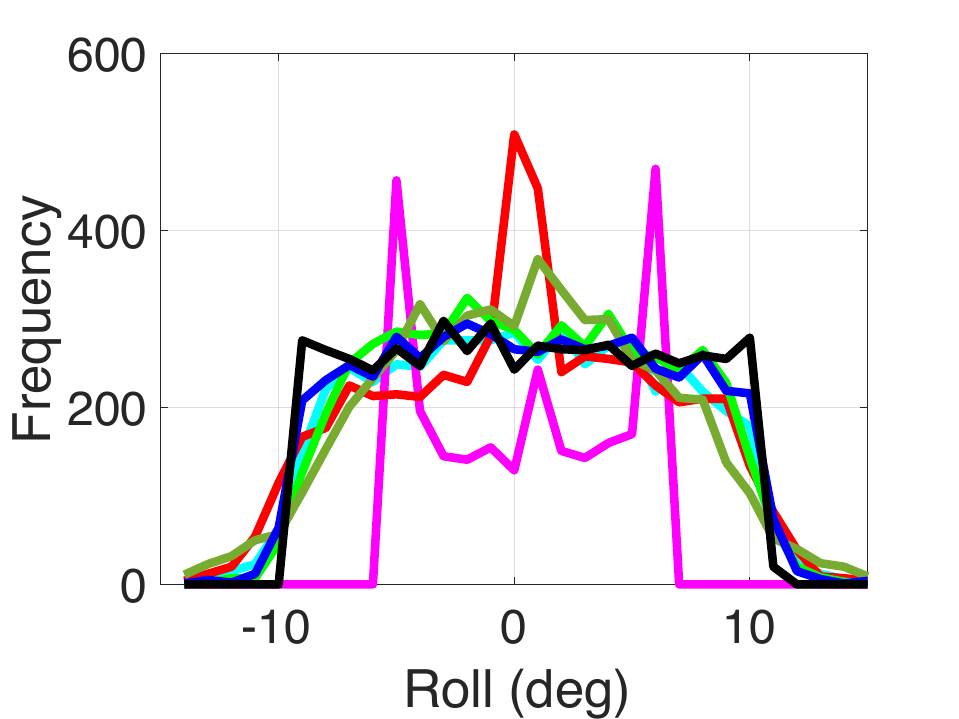}} &
		{\includegraphics[width=0.25\columnwidth]{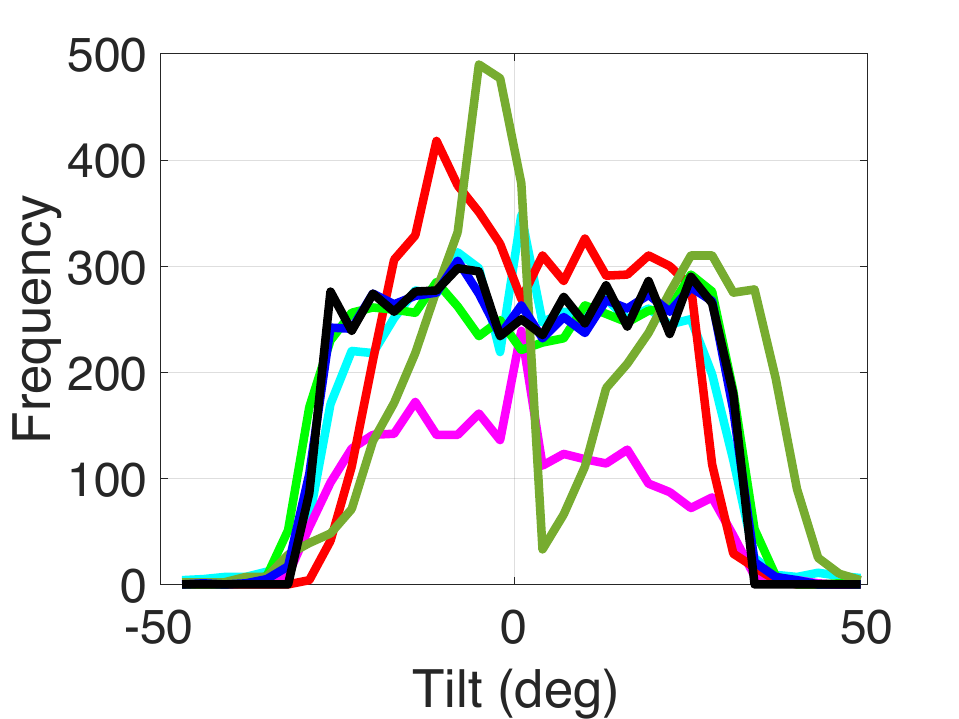}} &
		{\includegraphics[width=0.25\columnwidth]{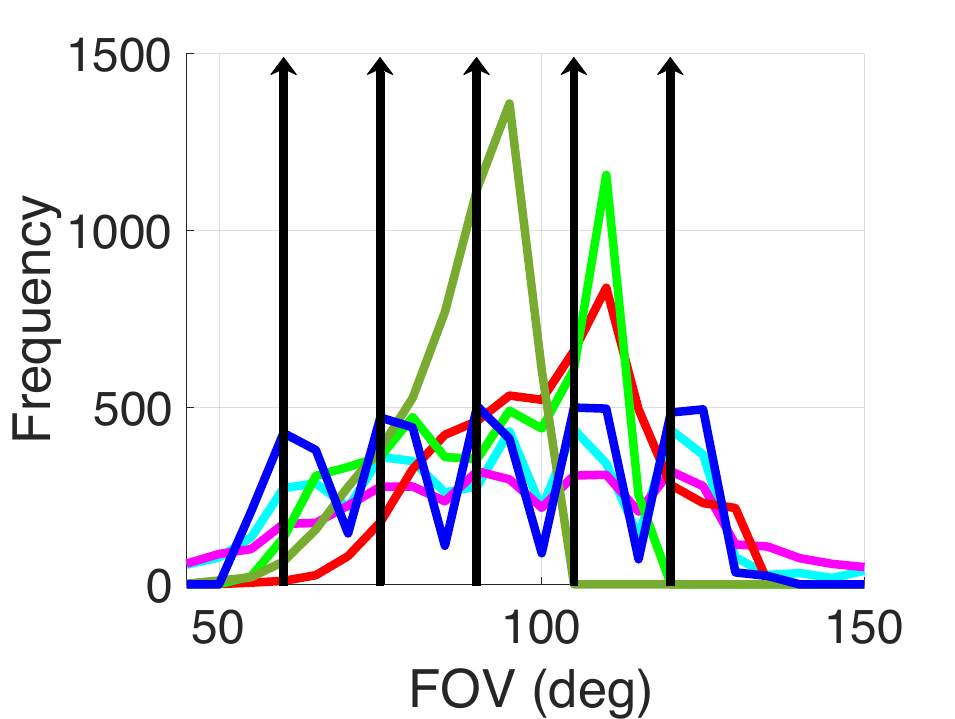}} 
	\end{tabular}
	\caption{Distribution of ground truth and estimated camera parameters for the PanoContext-{\em f}{\bf R} test sets.}
\vspace{-2em}
	\label{fig:sotaHist}
\end{figure}

 Fig. \ref{fig:sotaHist} compares distributions of estimated and ground truth parameters for the PanoContext-{\em f}{\bf R} dataset.  The distributions for camera rotation are generally reasonable, although the method of Simon et al.~\cite{simon2016simple} exhibits odd preferences for specific  roll and tilt angles,  the Hold-Geoffroy system~\cite{hold2018perceptual} has an overly strong bias to zero roll, and when trained on GSV, the CTRL-C system~\cite{lee2021ctrl} develops a bias against small upward tilts.

%

The sampling of five discrete FOVs in the PanoContext-{\em f}{\bf R} dataset allows us to visualize the sensitivity of the algorithms to focal length.  While the geometry-based algorithms all show modulation with the sampled FOVs (albeit weak for Simon et al~\cite{simon2016simple}), the deep learning methods do not, each forming a single broad peak.  Our {\em f}{\bf R} system appears to be best able to pick up the FOV signal from the data without a strong prior bias.  

The ability of the deep learning methods to estimate FOV/focal length might in part be cue to limitations in the range of FOVs in their training datasets.  However, in the supplementary material we show that this does not fully account for the superiority of our {\em f}{\bf R} system, which we believe derives from more direct geometry and probabilistic modeling.


%


\subsection{Predicting Reliability}
\label{sec:esterrresults}
We employ three global cues to predict camera parameter estimation error:  1) The minimum number of segments over the three Manhattan directions, 2) Entropy over our parameter grid search and 3) mean log likelihood of the final parameter estimate:  See supplementary material for a visualization of how parameter error varies as a function of these cues.  
To estimate the reliability of {\em f}{\bf R} estimates at inference, we use the PanoContext-{\em f}{\bf R} training set to fit a regression model that uses these three cues jointly to  predict the absolute error in each of the camera parameters:  We collect these cues into a 3-vector,  use the training data to compute a whitening transform and then use KNN regression in the 3D whitened space to estimate error, again selecting K by 5-fold cross-validation.  


Table \ref{tab:accumulated_mean} assesses the ability of this model to predict error on the held-out PanoContext-{\em f}{\bf R} test set.  The table shows the mean error obtained if we accept only the top 25\%, 50\%, 75\% or 100\% of estimates predicted to have least error.  The model is remarkably effective.  For example, by accepting only the 25\% of estimates predicted to be most reliable, mean error declines by 24\% for roll, 52\% for tilt and 66\% for focal length.  This ability to predict estimation error  can be extremely useful in applications where the camera can be recalibrated on sparse frames, or when conservative actions can be taken to mitigate risks. 

\begin{table}[htpb!]
	\caption{Parameter MAE for 25\%, 50\%, 75\% and 100\% of held-out PanoContext-{\em f}{\bf R} test images predicted to be most reliable.}
	\centering
	
	\begin{tabular}{|r|c|c|c|}
		\hline
		{\bf \% of images}  & \textbf{Roll MAE (deg)}& \textbf{Tilt MAE (deg)} & \textbf{Focal length MAE (\%)} \\ \hline 
		25\%   & \multicolumn{1}{c|}{0.56}    & 0.68 & \multicolumn{1}{c|}{2.8\%}          \\ \hline
		50\% & \multicolumn{1}{c|}{0.59}       & 0.75  &  \multicolumn{1}{c|}{3.9\%}            \\ \hline
		75\%   & \multicolumn{1}{c|}{0.63}       & 0.86   &  \multicolumn{1}{c|}{5.2\%}         \\ \hline
		100\%   & \multicolumn{1}{c|}{0.74}       & 1.42    &  \multicolumn{1}{c|}{8.3\%}       \\ \hline
	\end{tabular}
	\label{tab:accumulated_mean}
	\vspace{-2em}
\end{table}

\subsection{Run time}
\label{sec:runtime}
We ran our experiments on a 2.3GHz 8-Core Intel i9 CPU with 32 GB RAM and an NVIDIA Tesla V100-32GB GPU. Our {\em f}{\bf R} system as implemented is  slower than competing systems (Table \ref{tab:SOTA}), primarily due to our line segment detection system MCMLSD, which consumes on average 4.23 sec per image (Table \ref{tab:LSDPano}).

To address this issue, we have created a more efficient version of our system we call Fast-{\em f}{\bf R}, through two innovations.  First, we replace MCMLSD  with LSD, which consumes only 40 msec per image on average.  Second, we limit the number of iterations in our second search stage to 10.  This reduces average run time from 6.4 sec to 0.83 sec with only a modest decline in accuracy (Table \ref{tab:SOTA}).

\section{Limitations}
{\em f}{\bf R} relies on the Manhattan World assumption, i.e., that images contain aligned rectilinear structure.
The superior performance of {\em f}{\bf R} on the York Urban DB and PanoContext-{\em f}{\bf R} datasets, which are representative of common outdoor and indoor built environments, attests to the real-world applicability of this assumption.   
Nevertheless, the method will typically fail on scenes that do not conform to the Manhattan assumption.  This may include images with insufficient structure (see Supplementary Material for examples), nature scenes and non-Manhattan built
environments (e.g., the Guggenheim museums in New York or Bilbao).  Generalizing the {\em f}{\bf R} method to exploit additional scene regularities, e.g.,  Atlanta World~\cite{schindler2004atlanta} or quadrics, is an interesting direction for future research. 

Like all SOA methods reviewed and evaluated here, our {\em f}{\bf R} system assumes no non-linear distortions, a
central principal point, square pixels and zero skew.   These are reasonable approximations, since deviations from
these assumptions can be calibrated out in the factory or lab, and subsequent drift of these parameters is typically minor relative to changes in focal
length and rotation.  

While the Manhattan constraint can also be used to estimate the principal point, we find  that more accurate estimates of focal length and rotation are obtained by assuming a central principal point.  Simultaneous online estimation of focal length, principal point and camera rotation, as well as radial distortion, thus remains an interesting and challenging direction for future research.

	
\section{Conclusions}
We have employed existing and novel datasets to assess the reliability of online systems for joint estimation of focal length and camera rotation.  We show that the reliability of geometry-driven systems depends profoundly on the deviation measure employed and accurate probabilistic modeling.  Based on these findings, we have proposed a novel probabilistic geometry-driven approach called {\em f}{\bf R} that outperforms  four state-of-the-art competitors, including two geometry-based systems and two deep learning systems.  The {\em f}{\bf R} advantage is most pronounced for estimation of focal length, where the deep learning systems seem to struggle.  We note that the deep learning systems also do not predict pan angle, and so do not fully solve for the camera rotation.  We further demonstrate an ability to estimate the reliability of specific {\em f}{\bf R} predictions, leading to substantial gains in accuracy when systems can choose to reject parameter estimates judged to be unreliable.  We believe this will be useful in many real-world applications, particularly in mobile robotics and autonomous vehicles.

\section{Acknowledgement}
This work was supported by the University of Toronto, NSERC, the Ontario Research Fund, the York University VISTA and  Research Chair programs (Canada), and the Agency for Science, Technology and Research (A*STAR) under its RIE2020 Health and Biomedical Sciences (HBMS) Industry Alignment Fund Pre-Positioning (IAF-PP) Grant No. H20c6a0031 and AI3 HTPO Seed Fund (C211118014) (Singapore). 
\clearpage
%
%
\bibliographystyle{splncs04}
\bibliography{egbib}
\end{document}